%% file: cas-dc-sample.tex
\documentclass[a4paper]{cas-dc}

\usepackage[numbers]{natbib}
\usepackage[HTML]{xcolor}
\usepackage{booktabs}
\usepackage{multirow}
\usepackage{enumitem}
\setlist[itemize]{nosep}
\def\tsc#1{\csdef{#1}{\textsc{\lowercase{#1}}\xspace}}
\tsc{WGM}
\tsc{QE}
\tsc{EP}
\tsc{PMS}
\tsc{BEC}
\tsc{DE}

\begin{document}
\let\WriteBookmarks\relax
\def\floatpagepagefraction{1}
\def\textpagefraction{.001}

\title [mode = title]{From Open Waters to Enclosed Cabins: ProteusVPR for Cross-Scene Visual Place Recognition in Maritime Perception and Cabin Inspection}                      

\author[1,3]{Zexi Chen}[type=editor,
    auid=000,bioid=1]
\ead{chenzexi@sjtu.edu.cn}

\author[3]{Zitai Huang}
\ead{924000912020@sjtu.edu.cn}

\author[3,4]{Qiwen Gu}
\ead{2432178@tongji.edu.cn}

\author[2]{Zhiqi Li}
\ead{li.zhiqi1@coscoshipping.com}

\author[3]{Shengli Dong}
\ead{dong.shengli@coscoshipping.com}

\author[5]{Chenlei Wang}
\ead{wang.chenlei@coscoshipping.com}

\author[4]{Junqiao Zhao}
\ead{zhaojunqiao@tongji.edu.cn}

\author[1]{Hongdong Wang}
\ead{whd302@sjtu.edu.cn}

\author[2,5]{Bing Han}
\cormark[1]
\ead{han.bing@coscoshipping.com}

\affiliation[1]{organization={Shanghai Jiaotong University},
                city={Shanghai},
                country={China}}
\affiliation[2]{organization={COSCO SHIPPING Advanced Technology Institute},
                state={Shanghai},
                country={China}}
\affiliation[3]{organization={Shanghai Ship and Shipping Research Institute Co.LTD},
                state={Shanghai},
                country={China}}
\affiliation[4]{organization={Tongji University},
                city={Shanghai},
                country={China}}
\affiliation[5]{organization={Dalian Maritime University},
                city={Dalian, Liaoning},
                country={China}}

\cortext[cor1]{Corresponding author}
\fntext[fn1]{This work was supported in part by the National
Natural Science Foundation of China under Grant No.52502512 and in part by the Natural Science Foundation of Shanghai under Grant No.25ZR1402140.}

\begin{abstract}
Autonomous robotic inspection in maritime environments presents unique challenges for Visual Place Recognition (VPR) due to cross-scene perceptual shifts. Robots navigating ship-borne environments must transition between visually distinct domains: open decks with sparse textures and severe illumination changes, and enclosed cabins with repetitive structures and high visual ambiguity. Existing VPR methods, primarily designed for urban or indoor scenes, fail to generalize reliably across these starkly different scenarios. To address this, we propose ProteusVPR, a two-stage retrieval-refinement framework. The first stage employs any standard VPR model for initial image retrieval. The second stage introduces a geometric-visual estimation network that fuses the retrieved image with two temporally preceding frames, incorporating geometric descriptors, a local affine coordinate system, and camera azimuth encoding to achieve precise localization. To support this task, we introduce the XHZ dataset, an 8K-panoramic ship-borne dataset collected from an operational vessel, featuring multi-floor cabin structures, deck transition zones, and strict query-database separation for rigorous evaluation. Extensive experiments on the XHZ dataset demonstrate that ProteusVPR consistently improves localization accuracy across multiple VPR backbones, reducing mean localization error by over 60\% on average and that ProteusVPR offers an effective and robust solution for precise visual localization in challenging, cross-scene maritime environments. \textit{The full code will be released} \textbf{\textit{\href{https://github.com/erjieshizhe/pose-position}{here}}}.
\end{abstract}

\begin{keywords}
Visual Place Recognition \sep Intra-vessel Navigation \sep Perceptual Aliasing \sep Fine-grained Position Refinement
\end{keywords}

\maketitle

\section{Introduction}

Visual Place Recognition (VPR) aims to determine the location of a query image from visual observations and serves as an important component in autonomous driving, robotics, and augmented reality~\cite{li2025place}. Recently, with the growing interest in intelligent ship inspection and autonomous maritime systems, applying VPR to maritime environments has attracted increasing attention~\cite{thomas2022deep}.

\begin{figure*}[t]
\centering
\includegraphics[width=\textwidth]{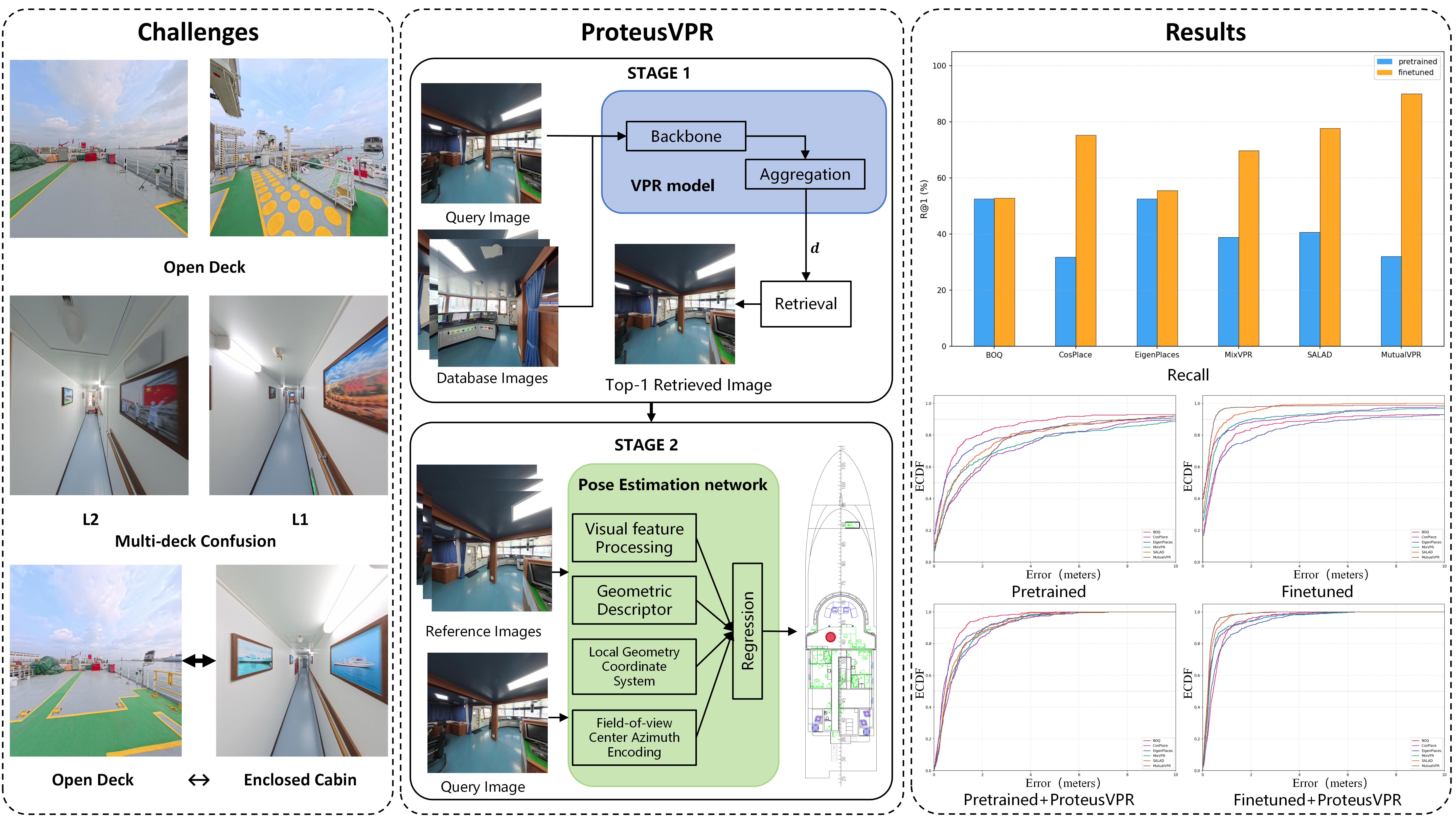}
\caption{Introduction of ProteusVPR. \textbf{Left:} Main challenges in maritime cross-scene VPR; \textbf{Middle:} architecture of the proposed retrieval refinement framework; \textbf{Right:} quantitative comparisons on the XHZ dataset.}
\label{fig:intro_guide}
\vspace{-10pt}
\end{figure*}

Compared with conventional urban or indoor scenarios, maritime environments introduce a more challenging cross-scene setting. In practical ship inspection tasks, the robot usually moves across two drastically different visual domains: open decks and enclosed cabins. Open-deck environments often contain sparse textures, repeated sky-water backgrounds, and severe illumination changes caused by weather and reflections. In contrast, enclosed cabins exhibit highly repetitive structures, narrow corridors, and multi-floor layouts with strong appearance similarity. Such a transition from open outdoor spaces to compact indoor environments poses significant challenges to both retrieval robustness and localization precision, as illustrated in Figure.~\ref{fig:intro_guide}.

Existing VPR methods mainly focus on global descriptor learning and image retrieval. Representative methods include NetVLAD~\cite{arandjelovic2016netvlad}, CosPlace~\cite{berton2022rethinking}, EigenPlaces~\cite{berton2023eigenplaces}, MixVPR~\cite{ali2023mixvpr}, and SALAD~\cite{izquierdo2024optimal}, which improve retrieval performance through feature aggregation, viewpoint modeling, or large-scale training strategies. Other methods further introduce local matching or re-ranking modules for fine-grained retrieval, such as Patch-NetVLAD~\cite{hausler2021patch}, TransVPR~\cite{wang2022transvpr}, and CricaVPR~\cite{lu2024cricavpr}. More recently, semantic-aware and vision-language based approaches~\cite{woo2024context,garg2024revisit,shen2025structvpr++} further improve scene understanding capability.

Despite these advances, directly applying existing VPR systems to maritime cabin environments remains difficult. One limitation of current retrieval-based methods is that they mainly optimize image-level retrieval performance instead of precise localization. In open-deck environments, sparse textures and drastic appearance changes often reduce retrieval robustness. Meanwhile, enclosed cabins contain highly repetitive structures, making different locations visually similar and causing severe ambiguity for global descriptors. Furthermore, simply fine-tuning existing models on maritime datasets is often insufficient, since the underlying feature representations are still designed for generic large-scale place retrieval rather than repetitive indoor localization.

We therefore ask the following question: \textit{Can a general refinement framework be built upon existing VPR systems to adapt them to maritime cabin environments while further improving localization precision?}

To address this problem, we propose \textbf{ProteusVPR}\footnote{Named after the shape-shifting sea god Proteus, highlighting adaptation across maritime visual domains.}, a second-stage retrieval refinement framework for maritime visual localization. Instead of redesigning the retrieval backbone, ProteusVPR is designed to work together with existing VPR models. Specifically, the first stage employs a general VPR model to retrieve the most relevant reference image, while the second stage introduces a geometric-visual estimation network to refine the retrieved location. By estimating relative poses under a local affine coordinate system, the proposed framework improves localization accuracy and partially alleviates retrieval ambiguity in repetitive cabin environments. \textbf{By bridging the gap between VPRs in open-water scenarios (e.g., autonomous berthing) and enclosed environments such as AR-based cabin inspection, we believe the proposed ProteusVPR can enhance perception across the full spectrum of maritime settings—true to its name, Proteus.}

The main contributions of this paper are summarized as follows:

\begin{enumerate}

\item \textbf{We propose ProteusVPR, a general second-stage retrieval refinement framework} that adapts existing VPR systems to maritime cabin environments. The proposed geometric-visual estimation module leverages image sequences to improves localization precision beyond nearest-neighbor retrieval and partially mitigates retrieval ambiguity in repetitive indoor scenes.

\item \textbf{We construct the XHZ dataset} containing ship cabins, multi-floor structures, and deck transition regions with strict query-database separation. The dataset provides a challenging benchmark for fine-grained maritime VPR and localization evaluation.

\item \textbf{Extensive experiments} on multiple benchmarks demonstrate that ProteusVPR consistently improves the localization accuracy and robustness of multiple existing VPR models in maritime cross-scene environments while maintaining good generalization capability, showing its capability of upgrading localization performence for future autonomous cabin inspection.

\end{enumerate}

\section{Related Work}

Visual place recognition (VPR) is formulated as an image retrieval task that determines the geographic location of a query image from visual information alone. Existing methods are broadly categorized into four groups according to their technical evolution: global descriptor aggregation, two-stage re-ranking, context and cross-image correlation, as well as semantic and multi-modal fusion.

\subsection{Global Descriptor Aggregation}

Early deep learning methods extract compact global descriptors from entire images. NetVLAD~\cite{arandjelovic2016netvlad} first introduced a differentiable formulation of the Vector of Locally Aggregated Descriptors (VLAD), enabling end-to-end training.

CosPlace~\cite{berton2022rethinking} reformulated large-scale geo-localization as a classification problem, eliminating the need for explicit negative mining. EigenPlaces~\cite{berton2023eigenplaces} further clusters scenes to explicitly model different viewpoints, improving robustness to view changes.

For feature aggregation, MixVPR~\cite{ali2023mixvpr} adopts a multi-level feature mixing strategy without explicit local feature extraction. SALAD~\cite{izquierdo2024optimal} re-formulates NetVLAD's soft assignment as an optimal transport problem and introduces a ``dustbin'' cluster to discard uninformative features. When combined with a DINOv2~\cite{oquab2023dinov2} backbone, SALAD achieves competitive performance.

Universal VPR solutions have also been explored. AnyLoc~\cite{keetha2023anyloc} demonstrates that off-the-shelf features from self-supervised models perform well across diverse environments without VPR-specific fine-tuning. MegaLoc~\cite{berton2025megaloc} fuses multiple existing methods, training techniques, and datasets into a single retrieval model, achieving strong results on VPR, landmark retrieval, and visual localization.

Most relevant to this work are BoQ~\cite{ali2024boq} and MutualVPR~\cite{gu2026mutualvpr}. BoQ introduces a bag of learnable global queries inspired by the BERT mechanism, using cross-attention to extract discriminative information from input features. MutualVPR addresses supervision inconsistency caused by viewpoint changes through a mutual learning framework where unsupervised view self-classification and descriptor learning co-evolve, improving robustness to view variations and occlusions.

\subsection{Two-Stage Re-ranking}

Two-stage strategies first retrieve a shortlist of candidate images using global descriptors, then refine the list with local features or geometric information.

Local-feature-based methods are typical in this category. Patch-NetVLAD~\cite{hausler2021patch} extracts multi-scale local patches from NetVLAD residuals and matches them for robustness to viewpoint and condition changes. TransVPR~\cite{wang2022transvpr} employs a Vision Transformer (ViT) to generate global descriptors via multi-head attention, and treats the filtered output tokens as local features for spatial matching re-ranking.

Unified Transformer frameworks for retrieval and re-ranking have gained attention. R$^2$Former~\cite{zhu2023r2former} designs a unified Transformer where the re-ranking branch jointly considers feature correlation, attention scores, and spatial coordinates, enabling end-to-end training. Pair-VPR~\cite{hausler2025pair} proposes a two-stage training pipeline: first pre-training a Transformer with pose-aware image sampling, then jointly learning global descriptors and a binary classifier. ProGEO~\cite{hu2024progeo} leverages CLIP to generate learnable text prompts and enhances visual features through image-text contrastive learning.

Furthermore, SelaVPR~\cite{lu2024towards} and EffoVPR~\cite{tzachor2024effovpr} explore better exploitation of features from pre-trained foundation models (especially DINOv2) for re-ranking. PRGS~\cite{zuo2025prgs} constructs a patch-to-region graph search, exploiting neighbourhood correspondences among candidate images to improve re-ranking.

\subsection{Context and Cross-Image Correlation}

Within-batch inter-image context is increasingly recognized as important for discriminative feature learning. CricaVPR~\cite{lu2024cricavpr} proposes cross-image correlation-aware representation learning, using self-attention to associate multiple images in a batch and incorporating multi-scale local information. UniPR-3D~\cite{deng2025unipr} adopts a 3D geometric perspective, designing a feature aggregator based on a visual geometry foundation model to encode multi-view 3D information for universal VPR.

For sequence-to-frame and temporal association, CaseVPR~\cite{li2025casevpr} introduces a hierarchical matching strategy consisting of coarse sequence retrieval and fine-grained sequence matching. IMVPR~\cite{cao2025imvpr} implicitly integrates a bird's-eye view (BEV) representation into multi-view VPR, using descriptor queries to encode 3D spatial positions for parallel fusion of multi-view features. Although VLAD-BuFF~\cite{khaliq2024vlad} focuses primarily on aggregation efficiency, its self-similarity driven feature discounting also models correlations among local patches.

\subsection{Semantic and Multi-Modal Fusion}

Integrating semantics or vision-language models into VPR is another direction. SegVLAD~\cite{garg2024revisit} decomposes an image into ``thing'' and ``stuff'' segments using the Segment Anything Model (SAM) and encodes or retrieves these segments instead of whole images, significantly improving robustness. StructVPR++~\cite{shen2025structvpr++} distils structural and semantic knowledge into RGB global representations via segmentation guidance, balancing global retrieval and re-ranking.

Vision-language models have recently introduced new paradigms for place recognition. The Context-Based Visual-Language Place Recognition method~\cite{woo2024context} proposes a training-free approach to handle scene variations by extracting pixel-wise embeddings via zero-shot, language-driven semantic segmentation, constructing semantic image descriptors.

For domain-specific scenarios, customized solutions have been developed. Zhou et al.~\cite{zhou2025visual} designed a method for coastal scenes that fuses semantic and sequential constraints, guiding the model to focus on landmarks via semantic constraints and resolving perceptual confusion via temporal constraints. MRS-VPR~\cite{yin2019mrs} achieves fast sequence matching through multi-resolution sampling, demonstrating high matching efficiency for long-term visual navigation.

\subsection{Summary and Positioning}

VPR has evolved from global descriptor learning to two-stage re-ranking, cross-image context modelling, and the incorporation of semantics and language models. MutualVPR~\cite{gu2026mutualvpr}, through its mutual learning framework that adaptively resolves supervision inconsistency, provides a foundation for this study.

However, none of the above methods are specifically designed for cross-scene maritime environments (from open decks to enclosed cabins). Their retrieval accuracy degrades significantly under extreme visual domain shifts. To address this gap, this paper proposes ProteusVPR. Building upon the first stage of MutualVPR, we introduce a two-stage estimation network that incorporates geometric descriptors, a local affine coordinate system, and azimuth encoding. The proposed method bridges high-recall retrieval and high-precision localization in complex ship interior scenarios. Subsequent experiments systematically validate its effectiveness on ship interior data, indoor corridors, and outdoor benchmarks.

\section{Methodology}
\label{sec:method}

\begin{figure*}[t]
\centering
\includegraphics[width=\textwidth]{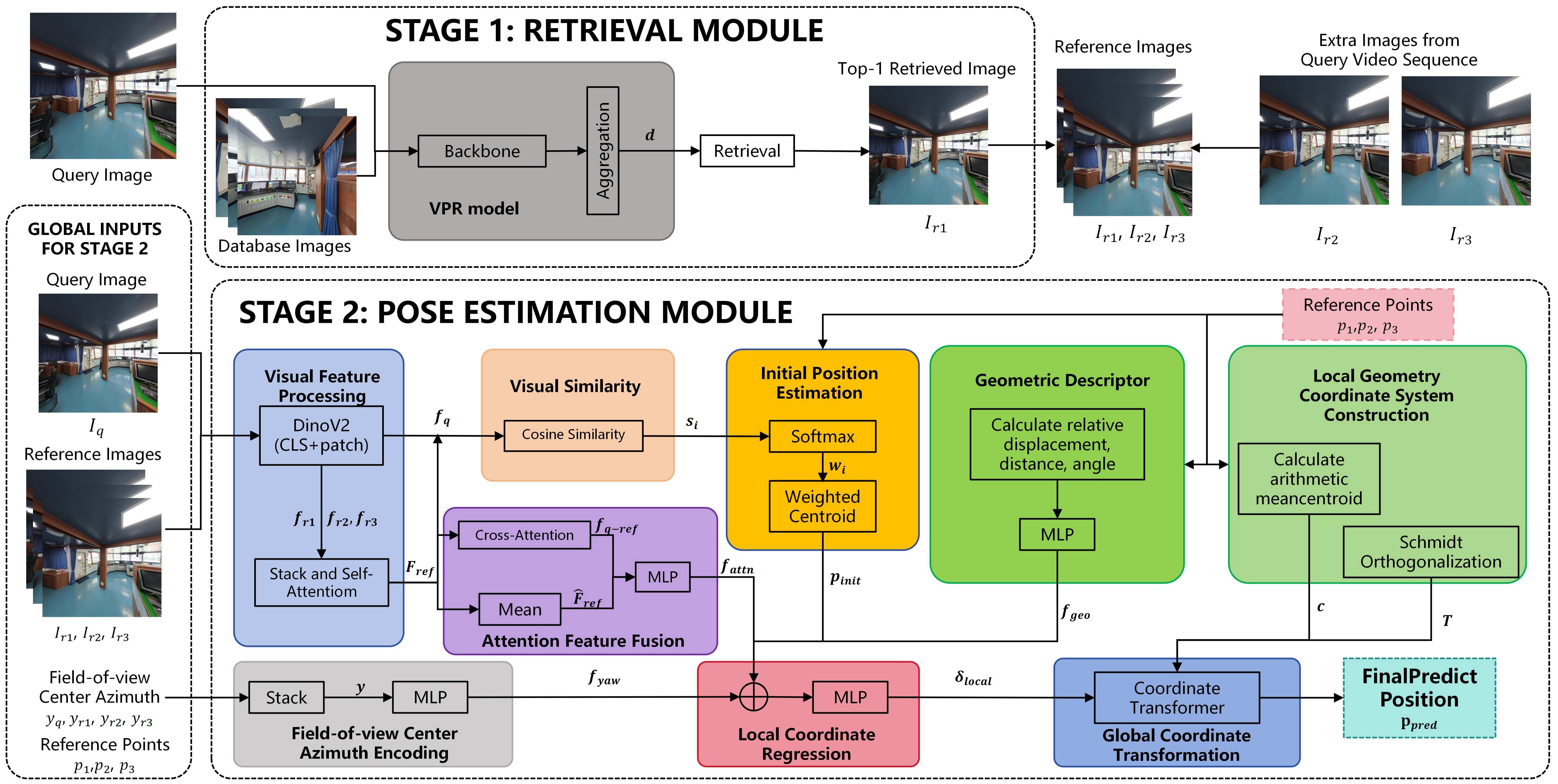}
\caption{Overall architecture of ProteusVPR. The first stage VPR model extracts a global descriptor for retrieval. The second stage fuses visual attention, geometric descriptors, a local affine coordinate system, and azimuth information to regress the query location.}
\label{fig:overall_architecture}
\vspace{-12pt}
\end{figure*}

The proposed method (shown in Figure.~\ref{fig:overall_architecture}) consists of two stages. The first stage is an image retrieval model that selects a reference image most similar to the query image from a database. The second stage takes this reference image together with the two preceding frames of the query image within its video sequence, and estimates the geographic location of the query image via a relative displacement estimation network. The two stages are detailed below.

\subsection{First Stage: Image Retrieval Model}

Given a query image \(I_q\), the first stage retrieves the most visually similar reference image \(I_{r1}\) from a large-scale geo-tagged image database, and outputs its geographic location \(\mathbf{p}_{r1}\) and field-of-view center azimuth \(y_{r1}\). This reference image is then used, along with the two preceding frames of the query image (denoted \(I_{r2}, I_{r3}\)), as input to the second stage.

The first stage adopts a visual place recognition (VPR) approach. In a standard VPR pipeline, an image is first encoded into a compact global descriptor, followed by nearest neighbor search in the descriptor space for retrieval.

\subsubsection{General Architecture of VPR}

A standard VPR model comprises a backbone network and an aggregation layer.

\paragraph{Backbone Network.}
The backbone takes an image \(I \in \mathbb{R}^{3 \times H \times W}\) as input and outputs a set of local feature maps. Typical backbones include convolutional neural networks or vision transformers.

\paragraph{Aggregation Layer.}
The local feature maps are aggregated into a single global descriptor \(\mathbf{d} \in \mathbb{R}^D\). This is commonly achieved via pooling strategies such as generalized mean (GeM) pooling or learnable modules like NetVLAD. The descriptor is usually L2-normalized to enable efficient similarity comparison.

\subsubsection{Retrieval Process}

Given a trained VPR model, all database images are pre-processed to extract their global descriptors \(\{\mathbf{d}_{\text{db}}^{(k)}\}\), which are then indexed for fast similarity search.

For a query image \(I_q\), the retrieval procedure is as follows:
\begin{enumerate}
    \item Extract its descriptor \(\mathbf{d}_q\) via a forward pass of the VPR model.
    \item Retrieve the most similar database image by maximum cosine similarity:
    \begin{equation}
    I_{r1} = \arg\max_{I_{\text{db}}^{(k)}} \cos(\mathbf{d}_q, \mathbf{d}_{\text{db}}^{(k)}),
    \end{equation}
    where \(\cos(\cdot, \cdot)\) denotes cosine similarity.
    \item Output \(I_{r1}\) together with its geographic location \(\mathbf{p}_{r1}\) and field-of-view center azimuth \(y_{r1}\).
\end{enumerate}

\subsection{Second Stage: Relative Displacement Estimation Network}

The second stage regresses the precise geographic location of the query image using three reference images: \(I_{r1}\) retrieved from the first stage, and \(I_{r2}, I_{r3}\), which are the two previous frames of the query image in its video sequence. The network adopts an end-to-end multi-modal fusion framework, as shown in Figure.~\ref{fig:overall_architecture}.

\subsubsection{Visual Feature Extraction and Processing}

All images (query image \(I_q\) and three reference images \(I_{r1}, I_{r2}, I_{r3}\)) share a fully frozen DINOv2 ViT-B/14 encoder. The input image size is adjusted to \(504 \times 504\). All pre-trained parameters remain frozen during training.

For each image, the encoder outputs a CLS token feature \(\mathbf{f}_{\text{cls}} \in \mathbb{R}^{768}\) and patch tokens \(\mathbf{F}_{\text{patch}} \in \mathbb{R}^{N_p \times 768}\), where \(N_p\) is the number of patches. To enhance the features, the CLS token is concatenated with the average of the patch tokens and then projected back to 768 dimensions via a linear projection layer \(\phi(\cdot)\):
\begin{equation}
\mathbf{f} = \phi\Big( \big[ \mathbf{f}_{\text{cls}};\; \frac{1}{N_p}\sum_{k=1}^{N_p} \mathbf{F}_{\text{patch}}^{(k)} \big] \Big) \label{eq:feature_enhance}
\end{equation}
where \([\,\cdot\,;\,\cdot\,]\) denotes concatenation. This yields the query feature \(\mathbf{f}_q\) and reference features \(\mathbf{f}_{r1}, \mathbf{f}_{r2}, \mathbf{f}_{r3}\).

Then, two layers of attention fusion are applied:
\begin{itemize}
    \item \textbf{Inter-reference attention:} a multi-head self-attention module is applied to the reference feature matrix \(\mathbf{F}_{\text{ref}} = [\mathbf{f}_{r1}, \mathbf{f}_{r2}, \mathbf{f}_{r3}]^\top \in \mathbb{R}^{3 \times 768}\), producing updated reference features \(\widehat{\mathbf{F}}_{\text{ref}} = [\widehat{\mathbf{f}}_{r1}, \widehat{\mathbf{f}}_{r2}, \widehat{\mathbf{f}}_{r3}]^\top\).
    \item \textbf{Query-reference attention:} the query feature \(\mathbf{f}_q\) is taken as the query, and \(\widehat{\mathbf{F}}_{\text{ref}}\) as the key and value, to compute cross-attention, yielding a query-relevant aggregated reference feature \(\mathbf{f}_{q\text{-ref}} \in \mathbb{R}^{768}\).
\end{itemize}

The mean of \(\widehat{\mathbf{F}}_{\text{ref}}\), i.e., \(\frac{1}{3}(\widehat{\mathbf{f}}_{r1} + \widehat{\mathbf{f}}_{r2} + \widehat{\mathbf{f}}_{r3})\), is concatenated with \(\mathbf{f}_{q\text{-ref}}\) and then encoded by an MLP into the final attention feature \(\mathbf{f}_{\text{att}} \in \mathbb{R}^{768}\).

Visual similarity weights are computed using the enhanced features:
\begin{equation}
w_{\text{sim}}^i = \frac{\exp(\tau \cdot \cos(\mathbf{f}_q, \mathbf{f}_{ri}))}{\sum_{j=1}^{3} \exp(\tau \cdot \cos(\mathbf{f}_q, \mathbf{f}_{rj}))}, \quad i=1,2,3 \label{eq:sim_weights}
\end{equation}
where \(\cos(\cdot,\cdot)\) is cosine similarity and \(\tau=10\) is a temperature coefficient. The three weights are denoted \(w_{r1}, w_{r2}, w_{r3}\) and form the vector \(\mathbf{w}_{\text{sim}} = [w_{r1}, w_{r2}, w_{r3}]^\top\), used for coarse location estimation.

\subsubsection{Coordinate Information Processing}

\paragraph{Geometric Descriptor Encoding}
For the geographic locations \(\mathbf{p}_{r1}, \mathbf{p}_{r2}, \mathbf{p}_{r3} \in \mathbb{R}^2\) of the three reference images, the following geometric quantities are computed for all pairs \((i,j)\) with \(i<j\):
\begin{align}
\boldsymbol{\Delta}_{ij} &= \mathbf{p}_{rj} - \mathbf{p}_{ri}, \label{eq:delta} \\
d_{ij} &= \|\boldsymbol{\Delta}_{ij}\|_2, \label{eq:distance} \\
\theta_{ij} &= \operatorname{atan2}(\Delta_{ij,y}, \Delta_{ij,x}) \label{eq:angle}
\end{align}

Each pair is represented by a 4‑dimensional vector \(\mathbf{g}_{ij} = [\Delta_{ij,x}, \Delta_{ij,y}, d_{ij}, \theta_{ij}]^\top\). It is encoded by an MLP \(\psi_{\text{pair}}\) into a 384‑dimensional feature; the encodings of all three pairs (total \(\binom{3}{2}=3\) pairs) are concatenated and fed into another MLP \(\psi_{\text{geo}}\) to obtain a global geometric descriptor \(\mathbf{f}_{\text{geo}} \in \mathbb{R}^{768}\):
\begin{equation}
\mathbf{f}_{\text{geo}} = \psi_{\text{geo}}\left( \big[ \psi_{\text{pair}}(\mathbf{g}_{12});\; \psi_{\text{pair}}(\mathbf{g}_{13});\; \psi_{\text{pair}}(\mathbf{g}_{23}) \big] \right) \label{eq:geo_descriptor}
\end{equation}

\paragraph{Local Affine Coordinate System Construction}
A 2D affine coordinate system is built using the three reference points:
\begin{align}
\mathbf{c} &= \frac{1}{3} \sum_{i=1}^{3} \mathbf{p}_{ri}, \label{eq:centroid} \\
\mathbf{v}_1 &= \mathbf{p}_{r2} - \mathbf{p}_{r1},\quad \mathbf{v}_2 = \mathbf{p}_{r3} - \mathbf{p}_{r1} \label{eq:basis_vectors}
\end{align}

Gram‑Schmidt orthogonalization is applied to \(\mathbf{v}_2\) with respect to \(\mathbf{v}_1\):
\begin{equation}
\mathbf{u}_2 = \mathbf{v}_2 - \frac{\mathbf{v}_2 \cdot \mathbf{v}_1}{\mathbf{v}_1 \cdot \mathbf{v}_1} \mathbf{v}_1 \label{eq:orthogonal}
\end{equation}

If the three points are collinear (\(\|\mathbf{u}_2\| < \epsilon\), where \(\epsilon\) is a very small positive number), set \(\mathbf{u}_2 = (-\mathbf{v}_{1,y}, \mathbf{v}_{1,x})^\top\).  
Normalization yields an orthonormal basis:
\begin{equation}
\mathbf{e}_x = \frac{\mathbf{v}_1}{\|\mathbf{v}_1\|},\quad \mathbf{e}_y = \frac{\mathbf{u}_2}{\|\mathbf{u}_2\|} \label{eq:orthonormal_basis}
\end{equation}

The transformation matrix is:
\begin{equation}
\mathbf{T} = [\mathbf{e}_x\ \mathbf{e}_y] \in \mathbb{R}^{2 \times 2} \label{eq:transform_matrix}
\end{equation}
This matrix transforms local coordinates to global coordinates: \(\mathbf{p}_{\text{global}} = \mathbf{c} + \mathbf{T} \cdot \mathbf{p}_{\text{local}}\).  
The coordinates of the reference points in the local coordinate system are:
\begin{equation}
\mathbf{p}_{ri}^{\text{local}} = \mathbf{T}^\top (\mathbf{p}_{ri} - \mathbf{c}), \quad i=1,2,3 \label{eq:local_coords}
\end{equation}

\subsubsection{Field-of-View Center Azimuth Encoding}

The field-of-view center azimuth \(y\) (also called camera boresight azimuth) is defined as the horizontal angle at which the cropped image is projected from the panoramic sphere. This angle describes the direction of the image's field-of-view center; it is not necessarily identical to the robot's own heading and may be interpreted as the installation angle of each camera relative to the robot's forward direction. In this paper, \(y\) is measured in degrees (see Section~\ref{sec:dataset}) and used directly as input to the neural networks. The azimuths of the query image \(I_q\) and reference images \(I_{r1}, I_{r2}, I_{r3}\) are denoted \(y_q, y_{r1}, y_{r2}, y_{r3}\), respectively.

The azimuths of the query and reference images are concatenated into a 4‑dimensional vector:
\begin{equation}
\mathbf{y} = [y_q,\ y_{r1},\ y_{r2},\ y_{r3}]^\top \in \mathbb{R}^4 \label{eq:yaw_vector}
\end{equation}

This vector is encoded by an MLP \(\psi_{\text{yaw}}\) into an azimuth feature:
\begin{equation}
\mathbf{f}_{\text{yaw}} = \psi_{\text{yaw}}(\mathbf{y}) \in \mathbb{R}^{128} \label{eq:yaw_features}
\end{equation}

\subsubsection{estimation and Coordinate Transformation}

A coarse location (weighted centroid) is first computed:
\begin{equation}
\mathbf{p}_{\text{init}} = \sum_{i=1}^{3} w_{ri} \cdot \mathbf{p}_{ri} \in \mathbb{R}^2 \label{eq:coarse_position}
\end{equation}
where the weights are given by Eq.~\eqref{eq:sim_weights}.

The attention feature \(\mathbf{f}_{\text{att}}\), geometric descriptor \(\mathbf{f}_{\text{geo}}\), coarse location \(\mathbf{p}_{\text{init}}\), and azimuth feature \(\mathbf{f}_{\text{yaw}}\) are concatenated and fed into a estimation MLP \(\psi_{\text{reg}}\) to output a displacement vector in the local coordinate system:
\begin{equation}
\boldsymbol{\delta}_{\text{local}} = \psi_{\text{reg}}\big( [\mathbf{f}_{\text{att}};\; \mathbf{f}_{\text{geo}};\; \mathbf{p}_{\text{init}};\; \mathbf{f}_{\text{yaw}}] \big) \in \mathbb{R}^2 \label{eq:local_displacement}
\end{equation}

Finally, the displacement is transformed to global coordinates:
\begin{equation}
\hat{\mathbf{p}}_q = \mathbf{c} + \mathbf{T} \cdot \boldsymbol{\delta}_{\text{local}} \label{eq:final_position}
\end{equation}
where \(\mathbf{c}\) and \(\mathbf{T}\) are defined by Eq.~\eqref{eq:centroid} and Eq.~\eqref{eq:transform_matrix}, respectively.

\subsubsection{Loss Function}

Before training the second stage, Z‑scale normalization is applied to the geographic coordinates of the entire dataset. For the 2D coordinates \((x, y)\) of all samples, the ranges along the \(x\) and \(y\) directions are computed, and the larger of the two is taken as the scaling factor \(s\); the minimum values \(\min_x\) and \(\min_y\) for each direction are also recorded. Each coordinate is then subtracted by the corresponding minimum, divided by \(s\), and mapped to the interval \([-1, 1]\). This normalization ensures equal scaling on both axes and constrains the entire dataset within \([-1, 1] \times [-1, 1]\).

During training, the estimation network directly predicts the position \(\hat{\mathbf{p}}_q^{\text{norm}}\) in the normalized coordinate space, and the loss is computed between this predicted normalized coordinate and the ground‑truth normalized coordinate. During inference, the normalized prediction is transformed back to real geographic coordinates via the inverse mapping:
\begin{equation}
\hat{\mathbf{p}}_q^{\text{real}} = \begin{bmatrix} \min_x \\ \min_y \end{bmatrix} + s \cdot \frac{\hat{\mathbf{p}}_q^{\text{norm}} + 1}{2},
\end{equation}
where \(\hat{\mathbf{p}}_q^{\text{norm}} \in [-1,1]^2\). All loss functions are computed in the normalized coordinate space.

The second stage is trained end‑to‑end with a loss function consisting of two terms. The total loss is:
\begin{equation}
\mathcal{L} = \lambda_{\text{abs}} \cdot \mathcal{L}_{\text{abs}} + \lambda_{\text{geo}} \cdot \mathcal{L}_{\text{geo\_cons}} \label{eq:total_loss}
\end{equation}
where \(\lambda_{\text{abs}}\) and \(\lambda_{\text{geo}}\) are hyperparameters balancing the loss terms.

\paragraph{Absolute Position Loss (\(\mathcal{L}_{\text{abs}}\)):} directly computes the Euclidean distance between the predicted position and the ground truth, without additional normalization.
\begin{equation}
\mathcal{L}_{\text{abs}} = \frac{1}{B} \sum_{b=1}^{B} \| \hat{\mathbf{p}}_q^{(b)} - \mathbf{p}_q^{\text{gt},(b)} \|_2 \label{eq:abs_loss}
\end{equation}
where \(B\) is the batch size and \(\mathbf{p}_q^{\text{gt}}\) is the ground‑truth geographic location of the query image. This loss drives the predicted position toward the ground truth.

\paragraph{Geometric Consistency Loss (\(\mathcal{L}_{\text{geo\_cons}}\)):} barycentric coordinate constraints ensure that the predicted position is reasonably represented by the reference points.
\begin{equation}
\begin{split}
\mathcal{L}_{\text{geo\_cons}} = \frac{1}{B} \sum_{b=1}^{B} \Bigg( \| \hat{\mathbf{p}}_q^{(b)} - \mathbf{c}^{(b)} \|_2 \quad & \\
+ \lambda_{\text{cons}} \cdot \max\Big(0,\; \min_{i \in \{1,2,3\}} \| \hat{\mathbf{p}}_q^{(b)} - \mathbf{p}_{ri}^{(b)} \|_2 - \rho \Big) &\Bigg) \label{eq:geo_cons_loss}
\end{split}
\end{equation}
where \(\mathbf{c}^{(b)}\) is the centroid of the reference points for the \(b\)-th sample (computed by Eq.~\eqref{eq:centroid}), \(\lambda_{\text{cons}}=0.3\) is the weight coefficient for the second term, and \(\rho=0.1\) is the distance threshold. The first term penalizes deviation of the predicted point from the centroid of the reference points; the second term adds an extra penalty when the predicted point is farther than \(\rho\) from all reference points, thereby enhancing geometric plausibility.

The second stage is trained independently and end-to-end. During training, the backbone parameters are frozen, and only the newly added attention modules, geometric encoder, azimuth encoder, and estimation head are updated.

\begin{figure*}[t]
    \centering
    \includegraphics[width=\textwidth]{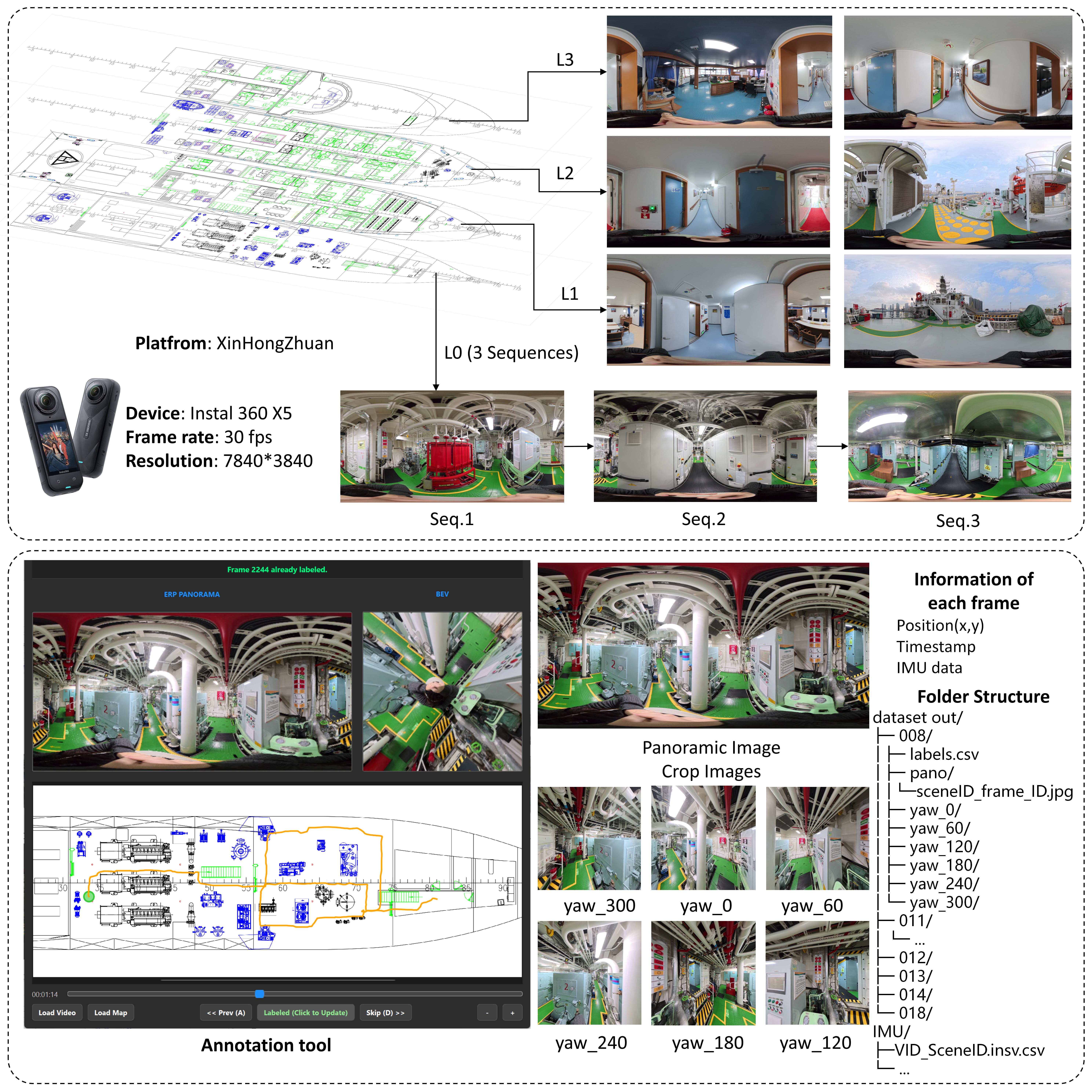}
    \caption{Overview of the XHZ dataset. The upper part shows the acquisition platform, device, and panorama examples; the lower part shows the annotation tool, exported images, annotations (position, timestamp, IMU), and folder structure.}
    \label{fig:dataset}
\vspace{-12pt}
\end{figure*}

\section{Dataset}
\label{sec:dataset}

\subsection{Overview}
To evaluate the proposed method on real-world ship inspection scenarios, the \textbf{Xinhongzhuan Visual Dataset} (demonstrated in Figure.~\ref{fig:dataset}, hereinafter referred to as the \textit{XHZ Dataset}) is introduced. This dataset was collected from an operational vessel, covering interior cabins and exterior deck areas to simulate typical ship inspection tasks. Its core characteristics include highly repetitive spatial layouts, strong visual similarity, constrained motion trajectories, and the presence of indoor-outdoor transition zones. These properties enable assessment of localization robustness when relying solely on visual cues.

The following subsections detail the acquisition, preprocessing, annotation protocols, and evaluation splits of the XHZ dataset.

\subsection{Data Acquisition}
\subsubsection{Hardware Configuration}
Data acquisition was performed using an Insta360 X5 professional panoramic camera, configured at a video resolution of $7840 \times 3840$ and a frame rate of 30 FPS. The device generates equirectangular images with a $360^\circ$ horizontal field of view and $180^\circ$ vertical field of view through dual-fisheye lens stitching, providing high-resolution source material for subsequent multi-view analysis.

\subsubsection{Acquisition Environment and Path Planning}
The acquisition area covers key compartments of the Xinhongzhuan ship, specifically including corridors in the living quarters, partial rooms, sections of the deck, and the engine room. To systematically cover heterogeneous spaces, the collection path was designed as follows:
\begin{itemize}
    \item Layered collection: the living area was divided into three floors (L1–L3), with each floor sampled independently.
    \item Segmented connection: the engine room (L0) was divided into three continuous stages based on functional layout, ensuring the start and end points of each stage were geographically connected to form a closed-loop topological verification ring.
\end{itemize}

\subsubsection{Acquisition Method}
To simulate the viewpoint height during real inspection or navigation tasks, the collector adopted a hand-held overhead approach and moved at a slow, steady pace. This method ensured a stable imaging perspective near eye level while minimizing motion blur.

\subsection{Image Preprocessing and Cropping}
Raw video streams were decoded into equirectangular panoramic images (resolution $7840 \times 3840$). To adapt to the input requirements of standard convolutional neural networks for planar perspective images, the following geometric transformation and cropping were applied:

\begin{enumerate}
    \item Reprojection: equirectangular images were converted into spherical projections based on spherical coordinate systems.
    \item Multi-view generation: using the field-of-view center azimuth (denoted as yaw) \(y\) as the baseline, starting from \(y = 0^\circ\), six directional perspective views were generated at \(60^\circ\) intervals. The horizontal field of view for each cropped image was set to \(120^\circ\). (Such yaw is elaborated upon in Section~\ref{sec:method}.)
    \item Overlap property: the theoretical overlap between adjacent cropped images is approximately 50\%, ensuring sufficient co-visible features between neighboring views for sequence matching and pose estimation.
\end{enumerate}

The final cropped images were uniformly scaled to a resolution of $2560 \times 2560$, preserving rich texture details.

\subsection{Ground Truth Annotation and Coordinate System}
\subsubsection{Annotation Tool and Method}
A dedicated annotation software was developed to improve efficiency. Its key features are as follows:
\begin{itemize}
    \item Temporal sampling: uniformly samples the original video at 5 Hz.
    \item View assistance: computes and displays a gravity-corrected front view in real time to aid orientation judgment.
    \item Automatic archiving: upon confirmation, the software automatically saves the current frame's raw panorama, six cropped sub-images, and corresponding coordinate labels.
\end{itemize}

\subsubsection{Coordinate Construction and Scaling}
Geographic coordinates were derived from CAD vector drawings of the ship. Operators clicked positions within the software interface; these points were then mapped to physical space by converting drawing units to meters (m) at a ratio of 80:1 based on the drawing's internal scale.

Annotation dimensions: each record contains the image filename, frame number, 2D planar coordinates \((x, y)\) (unit: m), and a millisecond-level timestamp.

\subsubsection{Coordinate Post-Processing}
Since different ship decks may overlap in 2D planar projections, a manual offset correction was applied to the initial coordinates. This step prevents VPR algorithms from incorrectly associating cross-floor scenes due to the lack of elevation constraints when calculating positive pairs:

\begin{itemize}
    \item L0: \(Y\)-coordinate \(-75\)
    \item L1: \(Y\)-coordinate \(-25\)
    \item L2: \(Y\)-coordinate \(+25\)
    \item L3: \(Y\)-coordinate \(+75\)
\end{itemize}

This operation significantly increases the coordinate difference between decks while preserving the relative topology within each floor, ensuring that evaluation metrics accurately reflect the model's ability to distinguish vertical levels.

\subsection{Dataset Versions and Scale}
Two incremental versions are released to accommodate research needs at different granularities:

\begin{itemize}
    \item Version 1 (manual): based on manual keyframe annotations at 5 Hz. Contains 2,996 panoramic images, resulting in a total of 17,976 cropped multi-angle images.
    \item Version 2 (interpolated): based on Version 1, linear interpolation was applied to unlabeled intermediate frames. Assuming uniform linear motion between manually annotated frames, coordinates for intermediate frames were derived by evenly distributing the displacement. This version expands to 17,982 panoramic images, with the total number of cropped images reaching 107,892 (overall 125,874 images).
\end{itemize}

Additional sensor data: raw IMU data sampled at 1000 Hz was recorded synchronously. Note that the panoramic video used for annotation underwent pre-processing electronic image stabilization (EIS); therefore, geometric deformation in the image frames does not perfectly align with the raw IMU timestamps. For precise kinematic analysis or sensor fusion using IMU data, strict spatiotemporal calibration is recommended to compensate for deviations introduced by stabilization.

\subsection{Data Structure and Partitioning}
\subsubsection{Storage Architecture}
The dataset is organized by scene. Each scene directory corresponds to a specific deck or area: \texttt{008} (L2), \texttt{011} (L1), \texttt{012}/\texttt{013}/\texttt{014} (three segments of L0 engine room), and \texttt{018} (L3). Within each scene directory, the following items are included:
\begin{itemize}
    \item \texttt{labels.csv}: centralized metadata file for all image annotations.
    \item \texttt{pano/}: original equirectangular panoramic images.
    \item \texttt{yaw\_0/}, ..., \texttt{yaw\_300/}: perspective images cropped at the specified yaw angles.
    \item Naming convention: image files follow the format \texttt{SceneID\_FrameID.jpg} to ensure unique traceability.
\end{itemize}

\subsubsection{Splitting Criterion}
Given the strong temporal correlation in sequential data, the dataset is strictly partitioned chronologically to avoid information leakage:
\begin{itemize}
    \item Training set: the first $2/3$ of frames in the sequence.
    \item Validation set: frames from $2/3$ to $5/6$ of the sequence.
    \item Test set: the last $1/6$ of frames in the sequence.
\end{itemize}

\subsubsection{Query-Database}
To simulate discrete observation and dense mapping scenarios in real-world VPR tasks, the following evaluation protocol is defined (as used in Section~\ref{sec:experiment}):
\begin{itemize}
    \item Validation: images with horizontal yaw angle $y = 0^\circ$ from each frame are selected as queries. The remaining yaw angles ($y = 60^\circ, 120^\circ, 180^\circ, 240^\circ, 300^\circ$) from the validation set constitute the database.
    \item Testing: similarly, images with $y = 0^\circ$ from each frame in the test set are selected as queries. The remaining yaw angles from the test set, combined with the training and validation sets, form the global database.
\end{itemize}

This protocol validates the model's cross-view localization ability when faced with a large number of visually similar scenes.

\section{Experiment}
\label{sec:experiment}

\subsection{Experimental Setup}

\subsubsection{Dataset and Evaluation}

Experiments are conducted on the \textbf{XHZ dataset}. The XHZ dataset (Section~\ref{sec:dataset}) utilizes the perspective images from Version~1 (manual). The data split, query/database construction, and coordinate design for distinguishing different decks follow the methods described in Section~\ref{sec:dataset}. Only perspective images are used throughout training, validation, and testing.

The following metrics are reported:

\begin{itemize}
    \item Recall: R@1, R@5 (\%). The positive sample threshold is a 2D Euclidean distance $\le 0.5$~m between images.
    \item Stage~1 error: Euclidean distance (in meters) when directly using the retrieved image coordinates as the estimate.
    \item Stage~2 error: Euclidean distance after refinement by the relative displacement estimation network.
\end{itemize}

Two complementary statistics are reported for the Stage~1 and Stage~2 errors:
\begin{itemize}
    \item \textbf{Mean Euclidean distance} (Table~\ref{tab:xhz_mean} in the Appendix): reflects average performance.
    \item \textbf{90\% quantile of Euclidean distance} (Table~\ref{tab:xhz_results} in the main paper): captures worst-case behavior and robustness against outliers.
\end{itemize}
Both statistics are reported for each deck level (L0 to L3) and for all baseline methods.

\subsubsection{Comparison Baselines}

Several VPR baselines are selected: BoQ~\cite{ali2024boq}, CosPlace~\cite{berton2022rethinking}, EigenPlace~\cite{berton2023eigenplaces}, MixVPR~\cite{ali2023mixvpr}, Salad~\cite{izquierdo2024optimal}, and MutualVPR~\cite{gu2026mutualvpr}. Each baseline is tested with two weight configurations:

\begin{itemize}
    \item \texttt{pre}: the publicly available pre-trained weights provided by the authors.
    \item \texttt{tra}: weights obtained after fine-tuning on the XHZ dataset.
\end{itemize}

\paragraph{Adaptation of BoQ output dimension.}
The original pre-trained weights of BoQ (based on DINO) have an output dimension of 12288, while all other baselines use 512 dimensions. For a fair comparison, a trainable linear projection layer is added at the end of BoQ to reduce its feature dimension to $512$. The projection layer is used as follows:
\begin{itemize}
    \item For the \texttt{pre} weights: only this projection layer is trained on XHZ.
    \item For the \texttt{tra} weights: during end-to-end fine-tuning on XHZ, the official unfreezing strategy of BoQ is followed, and the projection layer is updated simultaneously.
\end{itemize}
Thus, for both \texttt{pre} and \texttt{tra}, the projection layer is adapted to the XHZ dataset.

\subsubsection{Implementation Details}

Due to the default training settings of the compared baselines vary and the XHZ dataset has distinct characteristics (scene scale, number of images, inter-deck distribution), the fine-tuning hyperparameters for each baseline are adjusted according to the properties of XHZ. The hyperparameters listed below differ from the official defaults; parameters not mentioned remain at their official default values.

The hyperparameter changes for fine-tuning on XHZ (obtaining the \texttt{tra} weights) are as follows:
\begin{itemize}
    \item BoQ: \texttt{batch\_size=32, epochs=50, lr=3e-8}.
    \item CosPlace: \texttt{batch\_size=32, epochs=160, lr=1e-5, N=4, M=0.05, L=0.5, classifiers\_lr=0.01}.
    \item EigenPlaces: \texttt{batch\_size=32, epochs=50, M=0.05, N=2, lr=1e-7}.
    \item MixVPR: \texttt{batch\_size=32, epochs=50, lr=2e-5}.
    \item Salad: \texttt{batch\_size=32, epochs=50, lr=1e-4}.
    \item MutualVPR: \texttt{batch\_size=8, epochs=180, M=0.05, N=4, lr=1e-5, iterations\_per\_epoch=5000}.
\end{itemize}

For the \texttt{pre} weights (except BoQ), the official pre-trained models are used directly without any further parameter updates.  
For BoQ \texttt{pre}: before testing, the projection layer alone is trained on XHZ with the following hyperparameters: \texttt{batch\_size=32, max\_epochs=50, lr=6e-8, warmup\_epochs=10}. After this training, the BoQ model with the projection layer is used for evaluation.

The relative displacement estimation network (architecture described in Section~\ref{sec:method}) takes as input the retrieved images from the first stage, together with the two preceding frames in the video sequence of the query image, and outputs an accurate coordinate estimate.

Training hyperparameters for the relative displacement estimation network: during end-to-end training of the second-stage estimation network, the visual encoder (DINOv2) is kept frozen. Only the attention modules, geometric encoder, azimuth encoder, and regression head are updated. During training and validation, each sample is a 4-frame tuple with the same layer, same angle, temporally adjacent. In such a tuple, we pick the latest frame as the query, the other three as references. Batch size denotes the number of such tuples per batch. The hyperparameters are:
\begin{itemize}
    \item Batch size: 16
    \item Number of epochs: 300
    \item Initial learning rate: $2 \times 10^{-5}$
    \item Weight decay: $1 \times 10^{-4}$
    \item Learning rate scheduler: cosine annealing, warmup epochs = 30, minimum learning rate = $1 \times 10^{-7}$
    \item Loss weights for Eq.~\ref{eq:total_loss}: $\lambda_{\text{abs}} = 1.0$, $\lambda_{\text{geo}} = 0.2$
\end{itemize}

Hardware environment: all experiments are performed on a single NVIDIA RTX 4090 GPU (24GB VRAM) with PyTorch 2.6.0 + CUDA 12.4.

\begin{figure*}[!htbp]
    \centering
    \includegraphics[width=\linewidth]{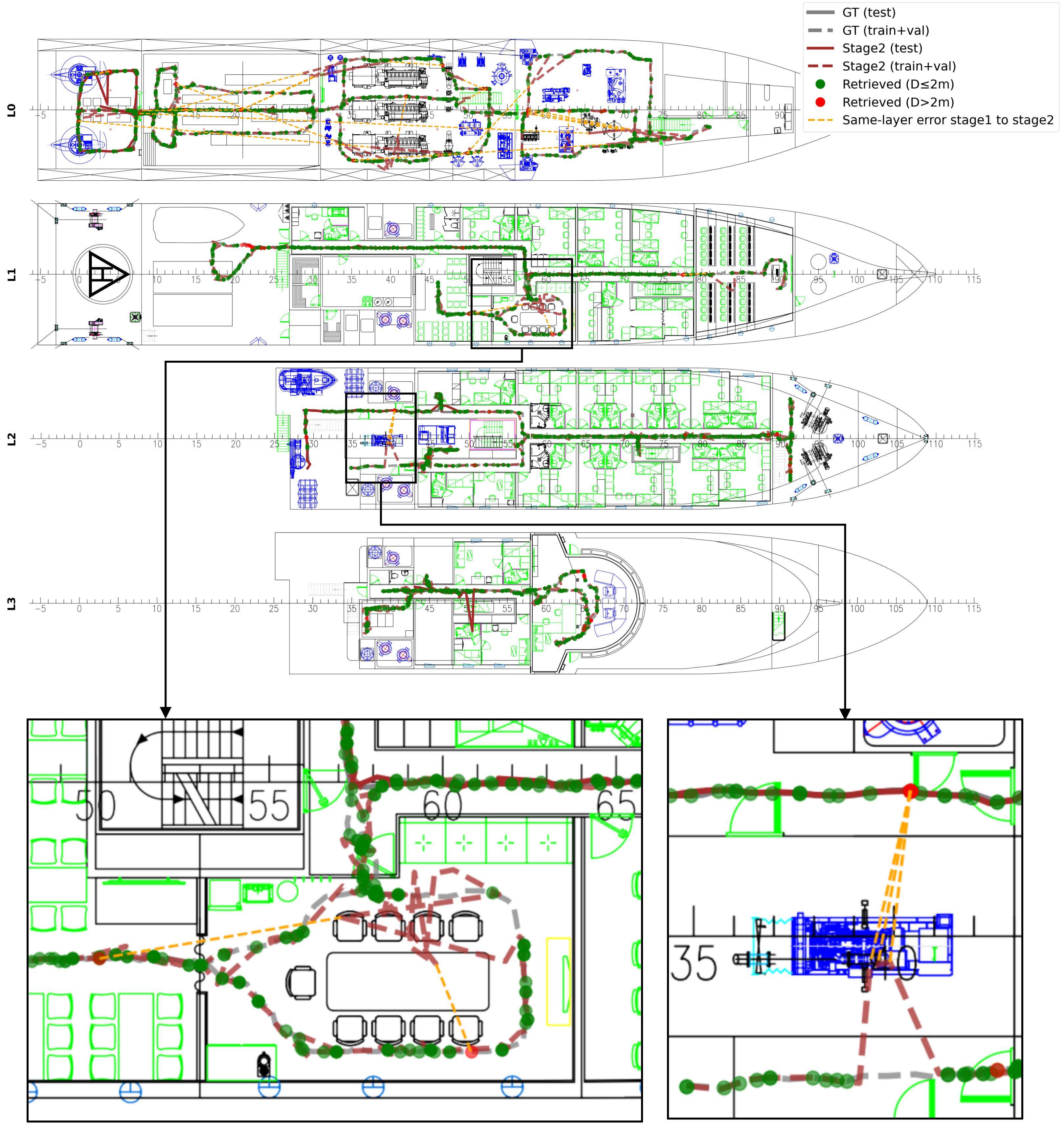}
    \caption{\textbf{Qualitative Demonstration} on the correction effect of the estimation module on intra-layer retrieval errors (retrieval module: fine-tuned MutualVPR). The connecting lines illustrate the correction relationship between a mis-retrieved point and its corresponding predicted point within the same layer. The two zoom-in panels below correspond to the boxed regions on L~1 (left) and L~2 (right), respectively.}
    \label{fig:trajectory_correction_intralayer}
    \vspace{-12pt}
\end{figure*}

\begin{figure*}[!htbp]
    \centering
    \includegraphics[width=\linewidth]{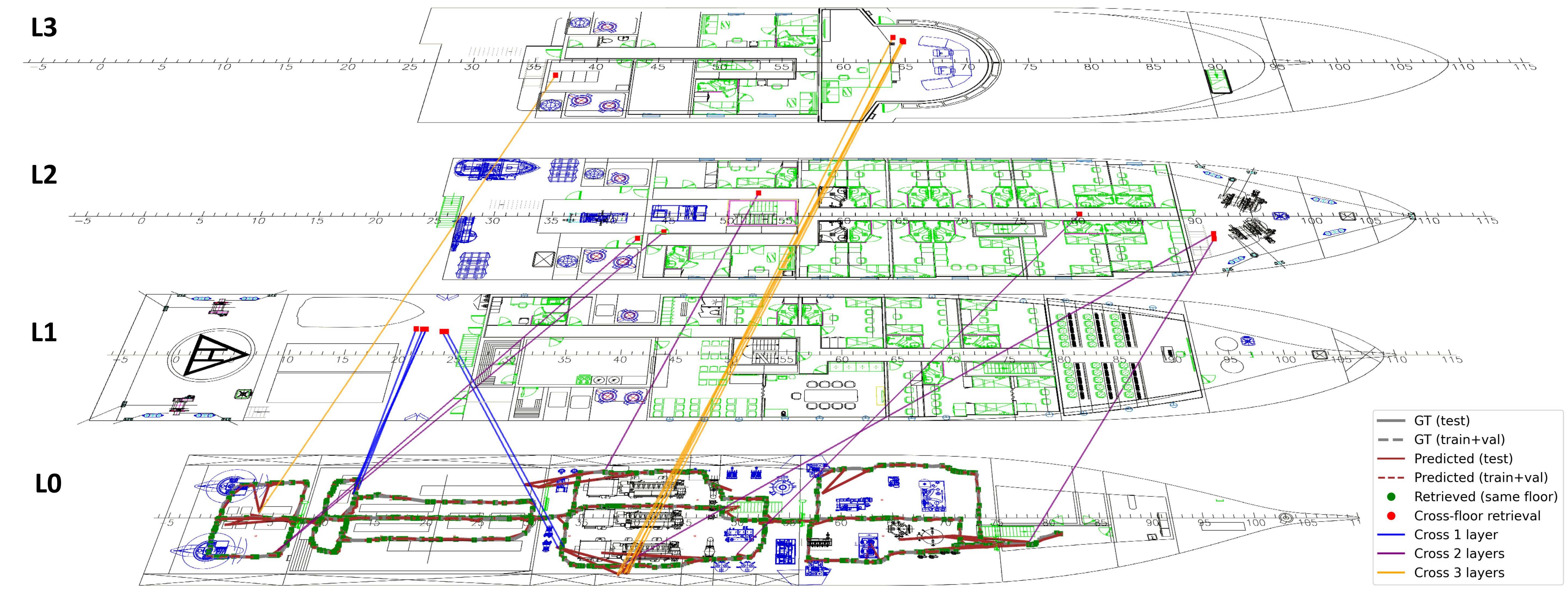}
    \caption{\textbf{Qualitative Demonstration} on the correction effect of the pose estimation on cross-layer retrieval errors (retrieval module: fine-tuned MutualVPR). In the case, we are locating query images in ''Layer 0'', The VPR might reach out to a mis-retrieved point (on a wrong layer) to its corresponding predicted point. Colors encode the number of layers crossed: blue for crossing one layer, purple for two layers, and orange for three layers. Refer to the legend for detailed marker and line styles.}
    \label{fig:trajectory_correction_crosslayer}
    \vspace{-10pt}
\end{figure*}

\subsection{Evaluation of the Proposed Dataset}

To validate the effectiveness of the proposed XHZ Dataset, all baselines (BoQ, CosPlace, EigenPlace, MixVPR, Salad, MutualVPR) are evaluated with both pre-trained (\texttt{pre}) and fine-tuned (\texttt{tra}) weights on the XHZ dataset in this section.
Table~\ref{tab:xhz_results} reports the 90\% quantile of Euclidean distance (in meters) for Stage~1 and Stage~2 errors, which reflects robustness against outliers. A qualitative demonstration is shown in Figure.~\ref{fig:trajectory_correction_intralayer}.
Table~\ref{tab:xhz_mean} reports the mean Euclidean distance, reflecting overall average performance.  
Both tables present results across layers 0 to 3. The analysis below jointly considers the two complementary metrics.

\input{table-XHZ.tex}      
\input{table-XHZ-mean.tex} 

\subsubsection{Overall Performance}

Each layer is analyzed independently, primarily based on R@1 and Stage~1 \textbf{90\% quantile errors} (reflecting worst-case robustness), while Stage~2 quantile errors are reported to validate the dataset's ability to support pose refinement. Mean errors are occasionally referenced as a complementary perspective.

\textbf{Layer 0.}  
MutualVPR (tra) achieves the best overall trade-off: R@1 = 88.4\%, Stage~1 quantile = 0.534~m, and its Stage~2 quantile is further reduced to 0.501~m. EigenPlaces (pre) has a lower Stage~1 quantile (0.899~m) but much lower recall (70.2\%). For MutualVPR (tra), the refinement from Stage~1 to Stage~2 (0.534 $\rightarrow$ 0.501~m) shows consistent improvement, validating that our dataset supports effective pose regression even under worst-case conditions.

\textbf{Layer 1.}  
MutualVPR (tra) attains R@1 = 90.9\%, Stage~1 quantile = 0.436~m, and Stage~2 quantile = 0.416~m. CosPlace (tra) yields R@1 = 72.7\% but a much larger Stage~1 quantile (7.040~m); SALAD (tra) achieves R@1 = 77.8\% and Stage~1 quantile = 0.785~m. In terms of mean error (supplementary), MutualVPR (tra) has 0.770~m (Stage~1) and 0.189~m (Stage~2), while SALAD (tra) shows 0.391~m and 0.241~m. The key observation is that the 90\% quantile improves from Stage~1 to Stage~2 for all strong methods, confirming the dataset's adequacy for robust pose refinement.

\textbf{Layer 2.}  
From Table~\ref{tab:xhz_results}, MutualVPR (tra) leads with R@1 = 87.6\%, Stage~1 quantile = 0.526~m, and Stage~2 quantile = 0.495~m. CosPlace (tra) follows (R@1 = 78.1\%, Stage~1 quantile = 1.204~m, Stage~2 quantile = 0.690~m). As a supplementary view, the mean errors in Table~\ref{tab:xhz_mean} show that SALAD (tra) has the smallest mean (Stage~1 = 0.556~m, Stage~2 = 0.364~m), while MutualVPR (tra) exhibits a larger mean (Stage~1 = 2.211~m) due to a long tail. Crucially, the quantile-based refinement for MutualVPR (tra) (0.526 $\rightarrow$ 0.495~m) demonstrates that our dataset helps tame worst-case outliers.

\textbf{Layer 3.}  
From Table~\ref{tab:xhz_results}, MutualVPR (tra) achieves the highest R@1 (97.3\%) and the smallest Stage~1 quantile (0.403~m), with Stage~2 quantile = 0.428~m --- a negligible increase given the already excellent error level. CosPlace (tra) ranks second (R@1 = 91.8\%, Stage~1 quantile = 0.429~m, Stage~2 quantile = 0.451~m). As a complementary metric, mean errors (Table~\ref{tab:xhz_mean}) show that SALAD (tra) has the smallest Stage~1 mean (0.274~m) and Stage~2 mean (0.242~m), but its recall is lower (R@1 = 82.2\%). MutualVPR (tra) shows a larger mean (Stage~1 = 1.519~m) due to a long tail, yet its Stage~2 mean plummets to 0.242~m.

In summary, MutualVPR (tra) delivers the highest recall and the best (or second-best) 90\% quantile errors in all layers. Across nearly all methods and layers, the pose refinement (Stage~2) consistently lowers the 90\% quantile error, often by a large margin. These results suggest that the XHZ dataset contains sufficient temporal and geometric structure to support effective coarse-to-fine localization: a strong retrieval model can achieve high recall, and the subsequent regression network can reliably reduce worst-case errors. The 90\% quantile serves as the primary robustness measure, while mean errors provide additional context.

\subsubsection{Effectiveness of Fine-Tuning on XHZ}

Fine-tuning on XHZ substantially improves most baselines. Taking layer~3 as an example:

- MutualVPR: R@1 rises from 41.1\% (\texttt{pre}) to 97.3\% (\texttt{tra}); Stage~1 quantile drops from 5.982~m to 0.403~m (93.3\% reduction); mean error drops from 9.917~m to 1.519~m (84.7\% reduction).  

- SALAD: R@1 from 47.9\% to 82.2\%; Stage~1 quantile from 3.501~m to 0.625~m (82.1\% reduction); mean error from 9.013~m to 0.274~m (97.0\% reduction).  

- CosPlace: R@1 from 47.9\% to 91.8\%; Stage~1 quantile from 2.755~m to 0.429~m (84.4\% reduction); mean error from 4.401~m to 4.416~m (0.3\% increase).

Overall, fine-tuning is essential for achieving high recall and low localization errors.

\textbf{Recall decrease despite error reduction.}  
Some models show a drop in R@1 after fine-tuning while Stage~1 errors become smaller. For instance, BoQ at layer~3: \texttt{pre}: R@1 = 61.6\%, Stage~1 = 1.274~m; \texttt{tra}: R@1 = 60.3\%, Stage~1 = 1.220~m.  
\(\Delta\)R@1 = -1.3\%, \(\Delta\)Stage1 = -0.054~m.  

Similarly, EigenPlaces at layer~0: \texttt{pre}: R@1 = 70.2\%, Stage~1 = 3.303~m; \texttt{tra}: R@1 = 68.9\%, Stage~1 = 2.606~m.  
\(\Delta\)R@1 = -1.3\%, \(\Delta\)Stage1 = -0.697~m.  

This behavior arises from the distinct natures of the two metrics. Recall is a binary indicator dependent on a fixed distance threshold (0.5~m), whereas Stage~1 error measures continuous Euclidean distance. Fine-tuning may push a few originally successful queries just beyond the threshold, slightly harming recall. Simultaneously, it reduces the distances of many previously far-off queries. The net effect is a leftward shift of the error distribution, leading to lower mean and quantile errors despite a minor recall decrease.

\subsection{Evaluation on ProteusVPR}

In this subsection, we primarily use Table~\ref{tab:xhz_mean} (mean errors) and Table~\ref{tab:xhz_results} (90\% quantile errors) to evaluate the pose estimation performance of ProteusVPR. For brevity, we refer to Table~\ref{tab:xhz_mean} as the ``mean error table'' and Table~\ref{tab:xhz_results} as the ``quantile table'' in the following discussion.

The proposed pose estimation (Stage~2) of ProteusVPR refines the initial retrieval coordinates using temporal and geometric information.

\textbf{Effectiveness of the pose estimation.}  
As shown in Table~\ref{tab:xhz_mean}, comparing Stage~1 and Stage~2 errors reveals consistent improvement for most models.    
Take layer~3:

- SALAD (tra): mean error \(0.274 \to 0.242\)~m (\(-11.7\%\)); 90\% quantile \(0.625 \to 0.472\)~m (\(-24.5\%\)).  

- MutualVPR (tra): mean error \(1.519 \to 0.242\)~m (\(-84.1\%\)); 90\% quantile \(0.403 \to 0.428\)~m (\(+0.025\)~m).  

- CosPlace (tra): mean error \(4.416 \to 0.275\)~m (\(-93.8\%\)); 90\% quantile \(0.429 \to 0.451\)~m (\(+0.022\)~m).

The pose estimation substantially reduces the mean error for all models, even when the quantile slightly increases. The small quantile increase occurs only when the Stage~1 quantile is already \(\leq 0.45\)~m and is negligible in practice.

\textbf{Impact of retrieval quality on regression accuracy.}  
Models with a long-tail error distribution (high mean but low quantile) benefit most from the pose estimation. MutualVPR (tra) has Stage~1 mean = 1.519~m and quantile = 0.403~m, indicating a few very large errors. The pose estimation reduces the mean by 84.1\%. In contrast, SALAD (tra) has a compact error distribution (mean = 0.274~m, quantile = 0.625~m), and the regression gain is smaller. Thus, the pose estimation is particularly effective when the first-stage retrieval has high recall but suffers from occasional catastrophic errors.

\subsubsection{Analysis of Pose Estimation Degradation Cases}
\label{subsubsec:blue_cases}

Two failure cases labled in blue are observed in Table~\ref{tab:xhz_results}. For CosPlace (tra) on Layer~3, the Stage~2 90\% quantile error (0.451~m) is slightly higher than the Stage~1 error (0.429~m). For MutualVPR (tra) on Layer~3, the Stage~2 90\% quantile error (0.428~m) is slightly higher than the Stage~1 error (0.403~m).

The pose estimation network predicts a displacement relative to the centroid of the three reference points (the Stage~1 retrieval result and its two preceding frames). When the Stage~1 retrieval is already accurate, the three reference points are spatially close. In this regime, the centroid is computed from points that lie near one another, and the network's robustness to small perturbations in the reference points decreases. Consequently, the predicted displacement may be more easily affected by localization noise. For CosPlace (tra) and MutualVPR (tra) on Layer~3, the Stage~1 retrieval error is relatively low (0.429~m and 0.403~m, respectively), meaning the three reference points are close together. In contrast, for entries where the Stage~1 error is larger, the reference points are more spread out. With a wider spatial distribution, the effect of small perturbations on the centroid is reduced, which helps the pose estimation reduce the error reliably. For example, for CosPlace (tra) on Layer~2, the Stage~1 90\% quantile error of 1.204~m is reduced to 0.690~m after pose estimation; for MutualVPR (tra) on Layer~0, from 0.534~m to 0.501~m; for SALAD (tra) on Layer~3, from 0.625~m to 0.472~m.

Except for these two cases, the Stage~2 90\% quantile errors are lower than the Stage~1 errors for all other rows and layers in Table~\ref{tab:xhz_results}. Furthermore, in terms of mean error (Table~\ref{tab:xhz_mean}), Stage~2 achieves lower values than Stage~1 for every row and layer, including the two blue-colored cases. These results collectively demonstrate that the pose estimation module remains effective overall.

\subsubsection{Case Study: Internal Mechanism of the Pose Estimation}

The internal mechanism of the pose estimation is explained using the XHZ dataset’s coordinate offset design and the architecture described in Section~\ref{sec:method}.

\textbf{Coordinate offsets between decks.}  
In XHZ, different decks (L0 to L3) are separated by fixed Y-axis offsets:  
L0: \(-75\), L1: \(-25\), L2: \(+25\), L3: \(+75\) (units: meters). This ensures that cross-floor errors produce large Euclidean distances, making them clearly detectable in evaluation.

\textbf{Inputs to the pose estimation.}  
The module takes three reference images: \(I_{r1}\) retrieved from Stage~1, and \(I_{r2}, I_{r3}\), which are the two immediately preceding frames of the query in its video sequence. All images are encoded by a shared frozen DINOv2 backbone. In the geometric branch, the three reference points are used to construct a centroid \(\mathbf{c}\) (Eq.~\ref{eq:centroid}) and a local affine coordinate system (Eqs.~\ref{eq:centroid}–\ref{eq:orthonormal_basis}). A coarse location \(\mathbf{p}_{\text{init}}\) is computed as a similarity-weighted centroid (Eq.~\ref{eq:coarse_position}). The regression head outputs a local displacement \(\boldsymbol{\delta}_{\text{local}}\) (Eq.~\ref{eq:local_displacement}) relative to the centroid, which is then transformed to global coordinates via Eq.~\ref{eq:final_position}.

\textbf{Correcting cross-floor retrieval errors.}  
When Stage~1 retrieves an image from a wrong layer (as demonstrated in Figure.~\ref{fig:trajectory_correction_crosslayer}), the retrieved coordinate deviates significantly from the true location due to the fixed Y-axis offsets between layers. However, the two preceding frames are taken from the same continuous trajectory as the query and therefore almost certainly belong to the correct layer. Consequently, the centroid \(\mathbf{c}\) of the three reference points is pulled toward the correct layer, reducing the initial error. The pose estimation network then predicts a small local offset to compensate for the remaining discrepancy. Additionally, the geometric descriptor captures the large distances between the off-layer point and the two correct ones, implicitly down-weighting the outlier. In this way, the module effectively mitigates cross-floor retrieval failures.

The combination of temporal context and the geometry-aware local coordinate system thus allows the pose estimation to correct severe retrieval failures, particularly those caused by layer confusion. This reduces the average localization error.

\subsection{Ablation Study}
\label{sec:ablation}

To verify the contribution of each key component in the proposed two-stage re-localization model, three ablation variants are designed based on MutualVPR (tra). All variants use the same training parameters as the full model and remove only specific modules from the network architecture.

Table~\ref{tab:ablation} reports the 90\% quantile localization errors (in meters) on the XHZ test set, reflecting robustness against outliers. Table~\ref{tab:ablation_mean} reports the mean Euclidean distance, reflecting overall average performance. The two tables jointly provide a view of each component's impact.

\subsubsection{Variant Definitions}
\begin{itemize}
    \item Full model: includes geometric descriptor encoding, local affine coordinate system construction and transformation, camera azimuth encoding, and the attention fusion module (i.e., the method described in Section~\ref{sec:method}).
    \item $w/o\;Geo$ – remove geometric descriptor encoding: the geometric descriptor encoding and its use are removed (i.e., Eq.~\ref{eq:geo_descriptor} and the associated MLP encoder are not computed). All other modules are kept.
    \item $w/o\;Loc$ – remove local coordinate system construction: the construction and transformation of the local affine coordinate system are removed (i.e., Eq.~\ref{eq:centroid} to Eq.~\ref{eq:orthonormal_basis} are not used). Geometric operations are performed in the global coordinate system: the estimation network directly outputs a global offset \(\hat{\mathbf{p}}_q = \mathbf{p}_{\text{init}} + \boldsymbol{\delta}_{\text{global}}\).
    \item $w/o\;Azi$ – remove camera azimuth encoding: the azimuth encoder and its use are removed (i.e., Eq.~\ref{eq:yaw_vector} and Eq.~\ref{eq:yaw_features} are omitted). The rest of the model remains unchanged.
\end{itemize}

\input{table-ablation.tex}      
\input{table-ablation-mean.tex} 

\subsubsection{Result Analysis}

Each variant is analyzed by examining both the 90\% quantile (Table~\ref{tab:ablation}) and the mean error (Table~\ref{tab:ablation_mean}).

\textbf{Geometric descriptor encoding (denoted as $w/o\;Geo$).}  
Removing the geometric descriptor encoding causes a large increase in localization error. For the 90\% quantile at layer~3, the error rises from \(0.432\)~m (full model) to \(2.974\)~m – a factor of approximately \(6.9\). The mean error at layer~3 increases from \(0.245\)~m to \(2.342\)~m (about \(9.6\) times). Similar degradation is observed across all layers. This indicates that the relative geometry among the three reference points (displacements, distances, and angles) provides an essential prior for regressing the query position.

\textbf{Local affine coordinate system (denoted as $w/o\;Loc$).}  
Removing the local coordinate system and performing estimation directly in the global coordinate system also degrades performance. At layer~3, the 90\% quantile increases from \(0.432\)~m to \(1.512\)~m, and the mean error from \(0.245\)~m to \(1.321\)~m. Similar patterns hold for other layers. This suggests that transforming geometric computations into a local affine coordinate system centred at the reference points provides rotation and translation invariance, simplifying the estimation task. Without this invariance, the network must learn both absolute positions and relative relationships in the global frame, leading to poorer generalization.

\textbf{Camera azimuth encoding (denoted as $w/o\;Azi$).}  
Removing the explicit azimuth encoder yields results close to the full model. For the 90\% quantile, the differences are within \(0.05\)~m (e.g., layer~2: \(0.537\)~m vs. \(0.498\)~m). The mean errors are slightly higher than the full model: at layer~2, \(0.318\)~m vs. \(0.285\)~m (\(+0.033\)~m); at layer~3, \(0.254\)~m vs. \(0.245\)~m (\(+0.009\)~m). Overall, the contribution of explicit azimuth encoding is limited. In the ship interior environment, the viewing direction (yaw) may already be implicitly present in the image content (e.g., wall textures on different sides of a corridor, door locations), allowing the network to infer azimuth from visual features alone.

Geometric descriptor encoding and the local affine coordinate system are core components of the two-stage estimation module; removing either leads to a substantial increase in both the mean and the 90\% quantile errors. Camera azimuth encoding provides only marginal benefits and can be considered optional. For practical applications, retaining the first two components is recommended to achieve the best localization accuracy.

\subsection{Summary of Experiments}

Experiments on the XHZ dataset yield the following quantitative findings.

\paragraph{Retrieval stage.}
Among all evaluated baselines, fine-tuned MutualVPR achieves the highest recall and the lowest 90\% quantile localization errors across all four layers. At layer~3, MutualVPR (tra) attains R@1 = 97.3\% and a Stage~1 90\% quantile error of 0.403~m, outperforming the next best baseline (CosPlace tra), which gives R@1 = 91.8\% and 0.429~m. The mean Stage~1 error of MutualVPR (tra) at layer~3 is 1.519~m, which is higher than that of SALAD (tra) (0.274~m), indicating a long‑tail error distribution. This tail is effectively addressed by the estimation module. By evaluating multiple well-known baselines on the proposed \textbf{\textit{XHZ Dataset}}, we reassure the effectiveness as well as reemphasize the importance of the dataset.

\paragraph{Pose Estimation stage.}
The relative displacement estimation network reduces the mean localization error substantially. At layer~3, the mean error of MutualVPR (tra) drops from 1.519~m (Stage~1) to 0.242~m (Stage~2), a reduction of 84.1\%. For SALAD (tra), the mean error decreases from 0.274~m to 0.242~m (–11.7\%). For CosPlace (tra), the mean error drops from 4.416~m to 0.275~m (–93.8\%). Averaged over all models and layers, the pose estimation reduces the mean localization error by 61.22\%.

\paragraph{Ablation study.}
Removing the geometric descriptor encoding ($w/o\;Geo$) increases the 90\% quantile error at layer~3 from 0.432~m to 2.974~m (factor of 6.9), and the mean error from 0.245~m to 2.342~m (factor of 9.6). Removing the local affine coordinate system ($w/o\;Loc$) raises the 90\% quantile to 1.512~m and the mean error to 1.321~m at layer~3. In contrast, removing camera azimuth encoding ($w/o\;Azi$) yields results close to the full model: at layer~2, the 90\% quantile differs by 0.039~m (0.537 vs. 0.498), and the mean error by 0.033~m (0.318 vs. 0.285); at layer~3, the mean error differs by 0.009~m (0.254 vs. 0.245).

These results demonstrate that the geometric descriptor and the local affine coordinate system are critical for achieving high localization accuracy in ship interior scenes, while explicit azimuth encoding offers only limited additional benefit.

\section{Conclusion}

This paper addresses the cross-scene visual place recognition problem in maritime environments, where a robot moves between open decks and enclosed cabins with drastically different visual characteristics. A two-stage retrieval refinement framework, ProteusVPR, is proposed. The first stage employs most of the common VPR models to retrieve a reference image from a geo-tagged database. The second stage introduces a geometric-visual estimation network that fuses the retrieved image with the two preceding frames of the query video sequence, incorporating geometric descriptors, a local affine coordinate system, and camera azimuth encoding to refine the query location.

Experimental results on the XHZ dataset demonstrate the effectiveness and robustness of the proposed framework for cabin-scale visual place recognition and localization. The proposed method consistently improves localization accuracy across different VPR backbones and ship decks, while the ablation studies further validate the importance of the geometric-aware design choices. Overall, the results show that the framework can effectively enhance existing VPR systems and provide reliable localization performance in challenging enclosed maritime environments.

The main contributions of this work are threefold. First, a general second-stage retrieval refinement framework, ProteusVPR, is proposed that adapts existing VPR systems to maritime cabin environments by improving localization precision beyond nearest-neighbor retrieval and partially mitigating retrieval ambiguity in repetitive indoor scenes. Second, the XHZ dataset is constructed, containing ship cabins, multi-floor structures, and deck transition regions with a strict query-database separation, providing a challenging benchmark for fine-grained maritime VPR and localization evaluation. Third, extensive experiments on multiple benchmarks demonstrate that ProteusVPR consistently improves the localization accuracy and robustness of several existing VPR models in maritime cross-scene environments.

\section{Limitations and Future Works}

Nevertheless, several limitations remain in the current study. i) The proposed refinement framework is inherently dependent on the quality of the initial retrieval stage; severe retrieval failures cannot be corrected during subsequent pose estimation. ii) The XHZ dataset was collected from a single vessel under relatively limited environmental variations, including lighting and operational conditions, leaving the generalization capability to other ship types and long-term deployments insufficiently explored. iii) The proposed estimation network assumes reliable reference coordinates, while its robustness to noisy odometry or imperfect SLAM inputs has not yet been systematically evaluated. Future work will therefore investigate uncertainty-aware retrieval and online adaptation mechanisms to improve robustness against retrieval outliers, as well as the construction of larger and more diverse maritime datasets to further enhance generalization performance.

\printcredits

\bibliographystyle{cas-model2-names}
 
\bibliography{cas-refs}


\end{document}

%% file: table-XHZ.tex
\begin{table*}
\centering
\footnotesize
\setlength{\tabcolsep}{3pt}
\caption{\textbf{Quantitative Demonstration} on the XHZ dataset (90\% quantile of Euclidean distance). ``pre'' denotes pre-trained weights, ``tra'' denotes fine-tuned weights. Metrics: R@1 and R@5 (\%); Stage1 and Stage2 errors (m) before and after the regression network. {\color[HTML]{FF0000} Red} indicates Stage2 enhanced the performance of Stage1 in the same row and layer while {\color[HTML]{0070C0} Blue} indicates the opposite (such corner cases will be discussed in Section~\ref{subsubsec:blue_cases}).}
\begin{tabular}{ll*{16}{c}}
\toprule
\multirow{2}{*}{Method} & \multirow{2}{*}{Mode} & \multicolumn{4}{c}{layer 0} & \multicolumn{4}{c}{layer 1} & \multicolumn{4}{c}{layer 2} & \multicolumn{4}{c}{layer 3} \\
 & & R@1 & R@5 & Stage1  & Stage2  & R@1 & R@5 & Stage1  & Stage2  & R@1 & R@5 & Stage1  & Stage2  & R@1 & R@5 & Stage1  & Stage2  \\
\midrule
\multirow{2}{*}{BoQ} & pre & 58.7 & 86.2 & 3.379 & {\color[HTML]{FF0000} 1.311} & 36.4 & 62.6 & 100.174 & {\color[HTML]{FF0000} 2.711} & 48.6 & 67.6 & 1.985 & {\color[HTML]{FF0000} 0.892} & 61.6 & 82.2 & 1.274 & {\color[HTML]{FF0000} 0.847} \\
                     & tra & 58.2 & 87.1 & 3.146 & {\color[HTML]{FF0000} 1.255} & 40.4 & 70.7 & 99.857 & {\color[HTML]{FF0000} 2.451} & 47.6 & 67.6 & 1.602 & {\color[HTML]{FF0000} 0.904} & 60.3 & 82.2 & 1.22 & {\color[HTML]{FF0000} 0.838} \\
\multirow{2}{*}{CosPlace} & pre & 32.9 & 65.3 & 4.528 & {\color[HTML]{FF0000} 1.87} & 9.1 & 25.3 & 19.613 & {\color[HTML]{FF0000} 4.778} & 40 & 62.9 & 50.137 & {\color[HTML]{FF0000} 1.199} & 47.9 & 75.3 & 2.755 & {\color[HTML]{FF0000} 1.026} \\
                     & tra & 69.8 & 91.6 & 4.286 & {\color[HTML]{FF0000} 1.643} & 72.7 & 93.9 & 7.04 & {\color[HTML]{FF0000} 2.128} & 78.1 & 93.3 & 1.204 & {\color[HTML]{FF0000} 0.69} & 91.8 & 98.6 & 0.429 & {\color[HTML]{0070C0} 0.451} \\
\multirow{2}{*}{EigenPlaces} & pre & 70.2 & 84 & 3.303 & {\color[HTML]{FF0000} 1.345} & 10.1 & 31.3 & 99.927 & {\color[HTML]{FF0000} 3.9} & 52.4 & 78.1 & 2.356 & {\color[HTML]{FF0000} 0.942} & 56.2 & 69.9 & 80.886 & {\color[HTML]{FF0000} 1.467} \\
                     & tra & 68.9 & 85.3 & 2.606 & {\color[HTML]{FF0000} 1.117} & 16.2 & 52.5 & 22.482 & {\color[HTML]{FF0000} 3.624} & 60 & 89.5 & 1.128 & {\color[HTML]{FF0000} 0.658} & 61.6 & 74 & 3.306 & {\color[HTML]{FF0000} 0.957} \\
\multirow{2}{*}{MixVPR} & pre & 46.7 & 76.9 & 3.973 & {\color[HTML]{FF0000} 1.595} & 4 & 12.1 & 50.021 & {\color[HTML]{FF0000} 3.977} & 43.8 & 71.4 & 31.629 & {\color[HTML]{FF0000} 1.269} & 54.8 & 64.4 & 51.141 & {\color[HTML]{FF0000} 1.614} \\
                     & tra & 73.8 & 88.9 & 1.631 & {\color[HTML]{FF0000} 0.841} & 61.6 & 94.9 & 13.748 & {\color[HTML]{FF0000} 3.341} & 65.7 & 92.4 & 0.842 & {\color[HTML]{FF0000} 0.496} & 74 & 93.2 & 1.044 & {\color[HTML]{FF0000} 0.657} \\
\multirow{2}{*}{SALAD} & pre & 50.2 & 78.2 & 3.151 & {\color[HTML]{FF0000} 1.376} & 11.1 & 23.2 & 23.477 & {\color[HTML]{FF0000} 3.621} & 42.9 & 69.5 & 2.274 & {\color[HTML]{FF0000} 0.971} & 47.9 & 67.1 & 3.501 & {\color[HTML]{FF0000} 1.45} \\
                     & tra & 80 & 92.4 & 1.247 & {\color[HTML]{FF0000} 0.723} & 77.8 & 98 & 0.785 & {\color[HTML]{FF0000} 0.428} & 69.5 & 94.3 & 1.269 & {\color[HTML]{FF0000} 0.716} & 82.2 & 97.3 & 0.625 & {\color[HTML]{FF0000} 0.472} \\
\multirow{2}{*}{MutualVPR} & pre & 36.4 & 64 & 4.368 & {\color[HTML]{FF0000} 1.729} & 5.1 & 18.2 & 21.761 & {\color[HTML]{FF0000} 4.412} & 41.9 & 68.6 & 3.178 & {\color[HTML]{FF0000} 1.08} & 41.1 & 58.9 & 5.982 & {\color[HTML]{FF0000} 1.659} \\
                     & tra & 88.4 & 97.8 & 0.534 & {\color[HTML]{FF0000} 0.501} & 90.9 & 99 & 0.436 & {\color[HTML]{FF0000} 0.416} & 87.6 & 95.2 & 0.526 & {\color[HTML]{FF0000} 0.495} & 97.3 & 100 & 0.403 & {\color[HTML]{0070C0} 0.428} \\
\bottomrule
\end{tabular}
\label{tab:xhz_results}
\end{table*}

%% file: table-XHZ-mean.tex
\begin{table*}[!tbp]
\centering
\footnotesize
\setlength{\tabcolsep}{3pt}
\caption{\textbf{Quantitative Demonstration} on the XHZ dataset (mean Euclidean distance). This table complements Table~\ref{tab:xhz_results} by reporting average errors.}
\begin{tabular}{ll*{16}{c}}
\toprule
\multirow{2}{*}{Method} & \multirow{2}{*}{Mode} & \multicolumn{4}{c}{layer 0} & \multicolumn{4}{c}{layer 1} & \multicolumn{4}{c}{layer 2} & \multicolumn{4}{c}{layer 3} \\
 & & R@1 & R@5 & Stage1  & Stage2  & R@1 & R@5 & Stage1  & Stage2  & R@1 & R@5 & Stage1  & Stage2  & R@1 & R@5 & Stage1  & Stage2  \\
\midrule
\multirow{2}{*}{BoQ} & pre & 58.7 & 86.2 & 0.911 & {\color[HTML]{FF0000} 0.572} & 36.4 & 62.6 & 24.621 & {\color[HTML]{FF0000} 0.901} & 48.6 & 67.6 & 0.76 & {\color[HTML]{FF0000} 0.517} & 61.6 & 82.2 & 1.887 & {\color[HTML]{FF0000} 0.456} \\
                     & tra & 58.2 & 87.1 & 1.441 & {\color[HTML]{FF0000} 0.586} & 40.4 & 70.7 & 23.723 & {\color[HTML]{FF0000} 0.811} & 47.6 & 67.6 & 0.724 & {\color[HTML]{FF0000} 0.507} & 60.3 & 82.2 & 1.832 & {\color[HTML]{FF0000} 0.437} \\
\multirow{2}{*}{CosPlace} & pre & 32.9 & 65.3 & 2.043 & {\color[HTML]{FF0000} 0.908} & 9.1 & 25.3 & 13.766 & {\color[HTML]{FF0000} 2.208} & 40 & 62.9 & 7.196 & {\color[HTML]{FF0000} 0.582} & 47.9 & 75.3 & 4.401 & {\color[HTML]{FF0000} 0.606} \\
                     & tra & 69.8 & 91.6 & 1.07 & {\color[HTML]{FF0000} 0.568} & 72.7 & 93.9 & 4.468 & {\color[HTML]{FF0000} 0.628} & 78.1 & 93.3 & 0.39 & {\color[HTML]{FF0000} 0.326} & 91.8 & 98.6 & 4.416 & {\color[HTML]{FF0000} 0.275} \\
\multirow{2}{*}{EigenPlaces} & pre & 70.2 & 84 & 0.899 & {\color[HTML]{FF0000} 0.524} & 10.1 & 31.3 & 18.009 & {\color[HTML]{FF0000} 1.737} & 52.4 & 78.1 & 3.549 & {\color[HTML]{FF0000} 0.484} & 56.2 & 69.9 & 11.597 & {\color[HTML]{FF0000} 0.638} \\
                     & tra & 68.9 & 85.3 & 0.801 & {\color[HTML]{FF0000} 0.477} & 16.2 & 52.5 & 12.891 & {\color[HTML]{FF0000} 1.568} & 60 & 89.5 & 2.442 & {\color[HTML]{FF0000} 0.395} & 61.6 & 74 & 4.739 & {\color[HTML]{FF0000} 0.5} \\
\multirow{2}{*}{MixVPR} & pre & 46.7 & 76.9 & 2.009 & {\color[HTML]{FF0000} 0.686} & 4 & 12.1 & 15.522 & {\color[HTML]{FF0000} 2.259} & 43.8 & 71.4 & 6.1 & {\color[HTML]{FF0000} 0.565} & 54.8 & 64.4 & 12.336 & {\color[HTML]{FF0000} 0.668} \\
                     & tra & 73.8 & 88.9 & 0.606 & {\color[HTML]{FF0000} 0.4} & 61.6 & 94.9 & 4.396 & {\color[HTML]{FF0000} 1.019} & 65.7 & 92.4 & 0.435 & {\color[HTML]{FF0000} 0.322} & 74 & 93.2 & 3.836 & {\color[HTML]{FF0000} 0.381} \\
\multirow{2}{*}{SALAD} & pre & 50.2 & 78.2 & 1.249 & {\color[HTML]{FF0000} 0.643} & 11.1 & 23.2 & 13.708 & {\color[HTML]{FF0000} 2.026} & 42.9 & 69.5 & 1.373 & {\color[HTML]{FF0000} 0.537} & 47.9 & 67.1 & 9.013 & {\color[HTML]{FF0000} 0.692} \\
                     & tra & 80 & 92.4 & 0.451 & {\color[HTML]{FF0000} 0.344} & 77.8 & 98 & 0.391 & {\color[HTML]{FF0000} 0.241} & 69.5 & 94.3 & 0.556 & {\color[HTML]{FF0000} 0.364} & 82.2 & 97.3 & 0.274 & {\color[HTML]{FF0000} 0.242} \\
\multirow{2}{*}{MutualVPR} & pre & 36.4 & 64 & 1.752 & {\color[HTML]{FF0000} 0.804} & 5.1 & 18.2 & 12.741 & {\color[HTML]{FF0000} 1.859} & 41.9 & 68.6 & 3.822 & {\color[HTML]{FF0000} 0.535} & 41.1 & 58.9 & 9.917 & {\color[HTML]{FF0000} 0.725} \\
                     & tra & 88.4 & 97.8 & 1.541 & {\color[HTML]{FF0000} 0.295} & 90.9 & 99 & 0.77 & {\color[HTML]{FF0000} 0.189} & 87.6 & 95.2 & 2.211 & {\color[HTML]{FF0000} 0.285} & 97.3 & 100 & 1.519 & {\color[HTML]{FF0000} 0.242} \\
\bottomrule
\end{tabular}
\label{tab:xhz_mean}
\end{table*}

%% file: table-ablation.tex
\begin{table}[t]
\centering
\caption{Ablation study (90\% quantile of Euclidean distance, in metres). $w/o\;Geo$: remove geometric descriptor encoding; $w/o\;Loc$: remove local affine coordinate system (global regression); $w/o\;Azi$: remove camera azimuth encoding.}
\begin{tabular}{lcccc}
\toprule
Model & layer 0 & layer 1 & layer 2 & layer 3 \\
\midrule
Full model & {\color[HTML]{FF0000}0.494} & {\color[HTML]{0070C0}0.412} & {\color[HTML]{FF0000}0.498} & {\color[HTML]{FF0000}0.432} \\
$w/o\;Geo$  & 2.826 & 3.093 & 2.913 & 2.974 \\
$w/o\;Loc$  & 1.205 & 1.732 & 1.525 & 1.512 \\
$w/o\;Azi$  & {\color[HTML]{0070C0}0.551} & {\color[HTML]{FF0000}0.409} & {\color[HTML]{0070C0}0.537} & {\color[HTML]{0070C0}0.479} \\
\bottomrule
\end{tabular}
\label{tab:ablation}
\vspace{-8pt}
\end{table}

%% file: table-ablation-mean.tex
\begin{table}[t]
\centering
\caption{Ablation study (mean Euclidean distance). This table complements Table~\ref{tab:ablation} by reporting average errors.}
\begin{tabular}{lcccc}
\toprule
Model & layer 0 & layer 1 & layer 2 & layer 3 \\
\midrule
Full model & {\color[HTML]{FF0000} 0.29} & {\color[HTML]{FF0000} 0.193} & {\color[HTML]{FF0000} 0.285} & {\color[HTML]{FF0000} 0.245} \\
$w/o\;Geo$  & 2.539 & 2.46 & 2.804 & 2.342 \\
$w/o\;Loc$  & 0.796 & 1.458 & 1.184 & 1.321 \\
$w/o\;Azi$  & {\color[HTML]{0070C0} 0.378} & {\color[HTML]{0070C0} 0.248} & {\color[HTML]{0070C0} 0.318} & {\color[HTML]{0070C0} 0.254} \\
\bottomrule
\end{tabular}
\label{tab:ablation_mean}
\vspace{-8pt}
\end{table}